\documentclass{article}
\usepackage{spconf,amsmath,graphicx}
\usepackage{balance}

\usepackage{amssymb}
\usepackage{hyperref}

\usepackage{etoolbox}  
\patchcmd{\thebibliography}{\leftmargin\labelwidth}
    {\itemsep 0.0pt \leftmargin\labelwidth}{}{}
    
\usepackage{easyReview}

\title{Multiscale IoU: A Metric for Evaluation of Salient Object Detection with Fine Structures}
\name{Azim Ahmadzadeh, 
        Dustin J. Kempton,
        Yang Chen, and
        Rafal A. Angryk}
\address{Department of Computer Science, Georgia State University, USA}

\begin{document}
%
\maketitle
\begin{abstract}
    
    General-purpose object-detection algorithms often dismiss the fine structure of detected objects. This can be traced back to how their proposed regions are evaluated. Our goal is to renegotiate the trade-off between the generality of these algorithms and their coarse detections. In this work, we present a new metric that is a marriage of a popular evaluation metric, namely Intersection over Union (IoU), and a geometrical concept, called fractal dimension. We propose Multiscale IoU (MIoU) which allows comparison between the detected and ground-truth regions at multiple resolution levels. Through several reproducible examples, we show that MIoU is indeed sensitive to the fine boundary structures which are completely overlooked by IoU and f1-score. We further examine the overall reliability of MIoU by comparing its distribution with that of IoU on synthetic and real-world datasets of objects. We intend this work to re-initiate exploration of new evaluation methods for object-detection algorithms.
\end{abstract}
%

%
%
\section{Introduction}\label{sec:introduction}
    The object-detection problem has been one of the primary targets of the computer vision field with a large variety of applications
    \cite{gavrila1999real,karpathy2015deep,xu2015show,bargoti2017deep, kupyn2018deblurgan, ahmadzadeh2019toward}.
    During the past decade, with the hardware's power catching up with the need of compute-intensive deep neural networks (and some other reasons \cite{lecun2019deep}) the computer vision field flourished at an unprecedented speed. In 2012, the classification performance exhibited by AlexNet \cite{krizhevsky2012imagenet} in the ImageNet Large Scale Visual Recognition Challenge (ILSVRC) \cite{ILSVRC15} outperformed humans and paved the road for more advanced algorithms \cite{alom2018history, zou2019object}. In the course of only six years (2012-2017) researchers managed to push the limits from highly accurate image classification to real-time localization of objects, and even better, to pixel-level region annotation (see \cite{zhao2019object} and the references therein). The achieved success has been made possible, at least partially, by the exuberant and popular competitions such as PASCAL VOC (2005-2012) \cite{everingham2010pascal}, ILSVRC (2010-2016), COCO (2015-present) \cite{lin2014microsoft}, and RVC (2018-present), and the excitement and directions they brought to the community.

    One side effect of such a fast growth, however, is the underlying assumption these general-purpose competitions impose to the object-detection task; that although the objective is to accurately localize (and classify) each object, a pixel-level precise detection is not of high priority. Each of the three components of a competition, i.e., dataset, ground-truth annotations, and evaluation metrics, enforces this assumption: (1) datasets often contain everyday objects (e.g., cars, pedestrians, ships, dogs), (2) objects are annotated coarsely (polygons used instead of drawing tools), and most importantly, (3) area-based metrics, e.g., Intersection over Union (IoU), are chosen for evaluating the algorithms. This intrinsic assumption is important for being able to easily rank the competing algorithms, as more complex methods may put many algorithms in gray areas. It is also critical that the chosen metric be able to handle the general purposes well, e.g., be effective even when only coarse annotations are available. That being said, as a consequence of such settings, the new algorithms manage to optimize their cost functions without a precise spatial estimate of objects, and become less sensitive to the fine boundary structures. Such an objective may not be relevant or even appropriate for many real-world problems. This realization is the first step in closing the gap between competitions' objective and the real challenges. We wish to contribute to this realization by introducing an alternative evaluation metric for general-purpose object-detection algorithms.
    
    \begin{figure*}[t]
        \centering\includegraphics[width=1.0\linewidth]{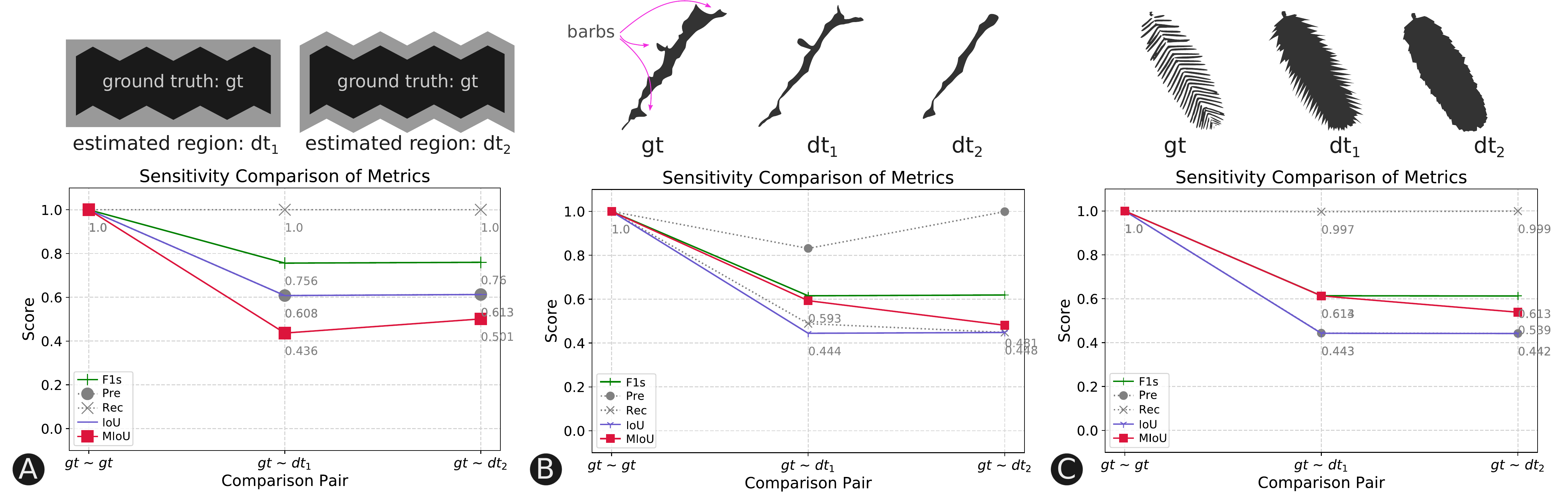}
        \caption{\footnotesize{Examples from different domains showing the metrics IoU, Precision, Recall, and F1-score fail to capture prominent differences between the proposed regions, $dt_1$ and $dt_2$, when compared with the ground-truth region, $gt$. Example A, depicts the issue in its simplest form. Example B and C illustrate the same issue using the mask of a solar filament and the mask of a leaf sample of the Metasequoia Glyptostroboides tree.}}
        \label{fig:3examples}
    \end{figure*}
    
    Many of the object-detection evaluation metrics are borrowed from the segmentation evaluation task. They either are very task-specific or have a large tolerance for the discrepancies between the ground-truth and detected regions' boundary. IoU is arguably the most popular measure in the second category. It quantifies the degree of which the ground-truth region is detected, i.e., intersection, relative to the area occupied by both of the ground-truth and detected regions, i.e., union. It is a simple, intuitive, and effective metric, but insensitive to details that should not be overlooked in many cases. In its simplest form, this undesirable tolerance is depicted in Fig. \ref{fig:3examples}-A, where two very different proposed regions, $dt_1$ and $dt_2$, are compared against a ground-truth region, $gt$. According to IoU (as well as Precision, Recall, and F1-score), $dt_1$ and $dt_2$ are equally good estimates for $gt$, notwithstanding the evident fundamental differences; $dt_2$ perfectly captures the jagged structure of $gt$'s boundary, whereas $dt_1$ only gives a bounding box for $gt$. We will introduce our proposed metric (MIoU) in Section \ref{sec:MIoU}, and explain how it can capture this discrepancy and puts $dt_2$ above $dt_1$ in its ranked list.
    
    IoU is just one of the metrics in the family of \textit{pixel-level errors}. Others examples include \textit{false-alarm} and \textit{missing-rate} pixel percentages \cite{huang1995quantitative} which are inspired by the concept of contingency table. Similarly, \textit{Precision} and \textit{Recall} are used in many studies, e.g., in benchmarking of image segmentation algorithms \cite{estrada2009benchmarking}. Whether computed on the regions' area or boundary, their binary view of `match or no-match' of pixels does not adequately quantify fine structural differences. \textit{Dice Similarity Coefficient} (DSC) is yet another area-based measure, that is only slightly different than IoU (by one intersection). But DSC easily approaches its upper bound \cite{zou2004statistical}, and to remedy this, \textit{Logit Transformation of DSC} (LTD) is often used instead. Inherently, LTD does not work well to quantify fine structural differences either. There are two other metrics which are also very popular, especially in medical image analyses, namely \textit{Global} and \textit{Local Consistency Errors} (GCE, LCE). They were used in preparation of the Berkeley Segmentation Dataset \cite{MartinFTM01}. But they are designed to judge between human-made segmentations, based on the needed refinements, which is automated in recent salient object detection algorithms, such as Mask R-CNN \cite{he2017mask}.
    
    There exist a number of other metrics that solely focus on the boundary structure of regions. A popular example of such class is \textit{LB\_Keogh Shape Indexing}, \cite{keogh2006lb_keogh}; a contour mapping measure that utilizes Dynamic Time Warping distance function. Despite its proven application, it cannot be used for evaluation of general-purpose object-detection algorithms because it disregards the area of objects, and moreover, it is a rotation invariant metric, which is not an appropriate assumption for all object detection problems. While many of these metrics have their strengths in specific use cases, we find two metrics most relevant to this study: IoU; because our metric is inspired by it, and F1-score; because it summarizes Precision and Recall which are the basics of any area-based metrics.

%
%
\section{Real-World Applications}\label{sec:realWorldApplications}

    In heliophysics, the spatial information of solar filaments can be used to determine the magnetic field orientation in a potentially associated coronal mass ejection (CME). This orientation understanding is critical, as this can help predict its impact on Earth's magnetic field and consequently our technology-dependant lives. The key information in the observed filaments is inferred from the angle of their `barbs' against filament's spine. The example illustrated in Fig. \ref{fig:3examples}-B shows a filament (captured by the Big Bear Solar Observatory\cite{denker1999synoptic} at `2012-02-21 19:12:50 (UTC)') and two proposed regions; one with the barbs and the other without any. If one were to evaluate the proposed regions with any of the previously listed metrics, only our proposed metric (MIoU) would provide a preference towards the example with barbs over the example with none.
    
    Our second example is from botany in the form of a plant species identification problem. It can be tedious and error-prone to use the classic method of manually parsing a (binary) tree of species, following a list of written features. Computer vision has made it much simpler these days to a degree that a mobile app can automatically extract the visual features of an arbitrary leaf sample and retrieve the most similar species \cite{kumar2012leafsnap}. One of the key features that is needed for construction of such a content-based image retrieval system is the leaves' shapes. Fig \ref{fig:3examples}-C gives an example of a leaf's shape, $gt$, (from LeafSnap dataset \cite{kumar2012leafsnap}, with ID `ny1041-04-1'), and two proposed shapes, $dt_1$ and $dt_2$. As the plot shows, similar to the previous examples, none of the listed metrics (except MIoU) differentiate between $dt_1$ and $dt_2$, and they fail to see the similarity that $dt_1$ exhibits to the ground-truth region, relative to $dt_2$.

%
%
\section{Multiscale IoU (MIoU)}\label{sec:MIoU}

    The object-detection evaluation metric that we propose is the marriage of two concepts: IoU and fractal dimension. The former is a similarity measure discussed in Section \ref{sec:introduction}. The latter is a classic measure that quantifies the complexity of fractals' structure and their lacunarity. In the following, we first review fractal dimension and a method for computing it, and then introduce our metric. 
    
    \noindent\textbf{Fractal Dimension and Box Counting Method.} Introduced in Fractal Geometry, fractal dimension gives a more general definition of `dimension', that quantifies the complexity of self-similar shapes, i.e., fractals. A number of different methods have been proposed to compute fractal dimension \cite{theiler1990estimating, mandelbrot1982fractal, barnsley1988science} among which \textit{box counting} is the most popular because it can be easily calculated on digital images. Using this method, fractal dimension ($D_{\text{Box}}(o)$) of an object $o$, can be calculated by the limit $\lim_{\delta \to 0} \frac{\log({n(o,\delta)})}{\log(1/ \delta)}$, where $\delta$ is the cell size of an evenly spaced grid, and $n(o,\delta)$ is the number of grid cells that overlap with the shape $o$. 
    In practice, the fractal dimension of the object $o$ is calculated in three steps: (1) superimpose $o$ on a grid of square cells of side length $\delta_i$. (2) For each $\delta_i$, using box counting method, count the number of grid cells that overlap with $o$ (or its contour). (3) Estimate the slope of the regression line of $\log(n(o,\delta_i))$ versus $\log(\delta_i)$, as $\delta_i$ decreases and produces finer grids, i.e., higher resolution. The estimated slope is the fractal dimension of $o$, that depending on the subject of study, can be computed on either its area or contour.

    \noindent\textbf{MIoU.} To fuse the multi-resolution concept of fractal dimension with IoU, we need a few definitions. Let $\Delta \subset \mathbb{N}$ be the set of all needed cell sizes, and $O$ be the set of all regions (of salient objects). Given an arbitrary region $o \in O$, and a cell size $\delta_i\!\in\!\Delta$, we define $s$, $s\!:\!O\!\times \mathbb{N}\!\to\!O$, to be a function that reduces the resolution of $o$ by replacing each $\delta_i$-by-$\delta_i$ cell with a single binary value $b$, and returns a new (lower resolution) region. The value of $b$ is 1, if its corresponding cell overlaps with $o$, and is 0, otherwise. For every $\delta_i$, this process can be carried out on both of the ground-truth ($o$) and detected ($\Tilde{o}\!\in\!O$) regions. Furthermore, let $n$, $n\!:\!O\!\to\!\mathbb{N}$, be another function that simply counts the number of cells a region spans over. This is equivalent to the number of pixels that form the given region after it is downsampled by $s$. Therefore, it does not depend on $\delta_i$.
        
    With the above tools, we can now define the \textit{intersection ratio} denoted by $r$, $r\!:\!O^2\!\times\!\Delta\!\to\![0,1]$, as shown in Eq. \ref{eq:ratio}. For a given ground-truth region $o$, a detected region $\Tilde{o}$, and a cell size $\delta_i$, $r$ measures the ratio of the number of cells $o$ and $\Tilde{o}$ have in common, over the number of cells they should have in common if they perfectly align.
        
    \begin{equation}\label{eq:ratio}
        r(o, \Tilde{o}, \delta_i) = \frac{n\big(s(o, \delta_i) \cap s(\Tilde{o}, \delta_i)\big)}{n\big(s(o, \delta_i)\big)}
    \end{equation}
    
    Note that for a given pair of regions, $r$ is a function of cell size, $\delta_i$. That is, intersection ratio can measure the alignment at any desired resolution level determined by $\delta_i$. In other words, if two regions of interest are well-aligned, i.e., their area and boundary pixels almost perfectly match, their subtle miss-alignment can still be captured in higher resolution levels. And if they are not well-aligned, either spatially or structurally, their slight alignments (if at all) can still be captured in lower resolution levels. Therefore, a proper similarity assessment can be made with a multiscale comparison. This observation explains our final step in defining MIoU. Given two detected regions, $\Tilde{o}$ and $\Tilde{o}'$, for the ground-truth region $o$, assuming that $\Tilde{o}$ is a much better estimate for $o$, than $\Tilde{o}'$, it is expected that, on average, $r(o, \Tilde{o}, \delta_i) \ge r(o, \Tilde{o}', \delta_i)$, for all $\delta_i\!\in\!\Delta$. Therefore, we propose the area under the curve of $r(o, \Tilde{o}, \delta)$ for all $\delta \in \Delta$ as a measure of alignment for two arbitrary regions, $o$ and $\Tilde{o}$. This can be formulated by the integral $MIoU(o, \Tilde{o}) = \int_{0}^{1} r(o, \Tilde{o}, \delta)\: d\delta$. By choosing $d\delta = \frac{1}{|\Delta|-1}$ and transforming $\delta$ to the range of $[0,1]$, the metric MIoU will also be limited to the interval $[0,1]$, where 1 implies the perfect alignment of $o$ and $\Tilde{o}$, and 0 indicates the opposite.
    
    Although MIoU can technically be computed on either of the regions' areas or boundaries, we find the use of boundaries more appropriate. This is simply due to the fact that objects' area grows faster than their perimeter. Therefore, the dissimilarities between objects' boundaries can be overshadowed by the large number of pixels their areas span over. This renders the intersection ratio $r$ ineffective. To avoid this, we only take into account the contour of objects in all of the above-mentioned definitions. Our implementation of MIoU, as well as all experiments in Section \ref{sec:experiments}, are made publicly available\footnote{Code: \href{https://bitbucket.org/gsudmlab/multiscale_iou/}{https://bitbucket.org/gsudmlab/multiscale\_iou/}}.

%
%
\section{Experiments and Results}\label{sec:experiments}
    
    \begin{figure}[t]
        \centering\includegraphics[width=1.0\linewidth]{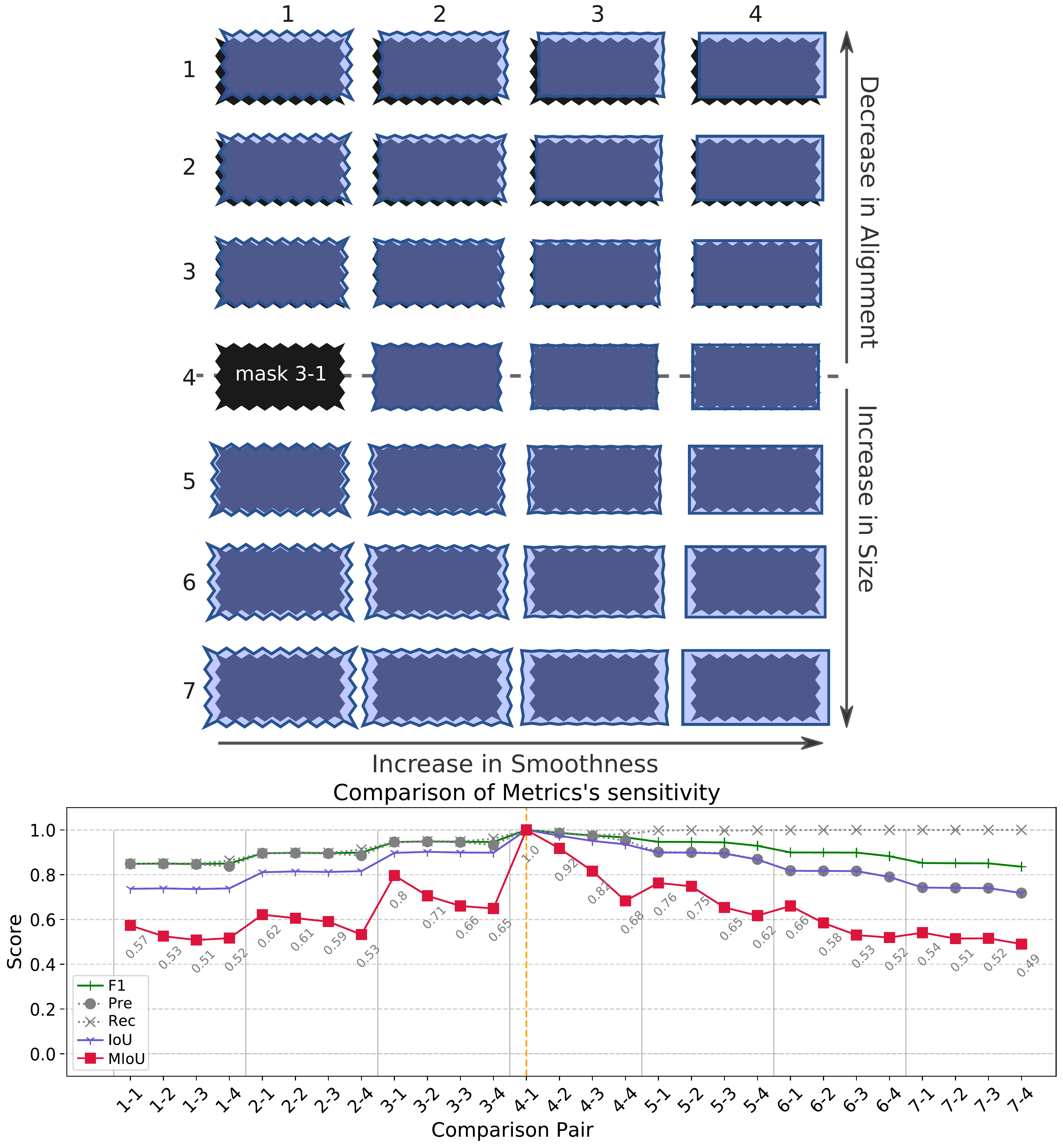}
        \caption{\footnotesize{Comparison of area-based metrics on the 28 estimates (shown on top) for the ground-truth region, indexed 3-1.}}
        \label{fig:example_f}
    \end{figure}
    
    \begin{figure}[t]
        \centering\includegraphics[width=1.0\linewidth]{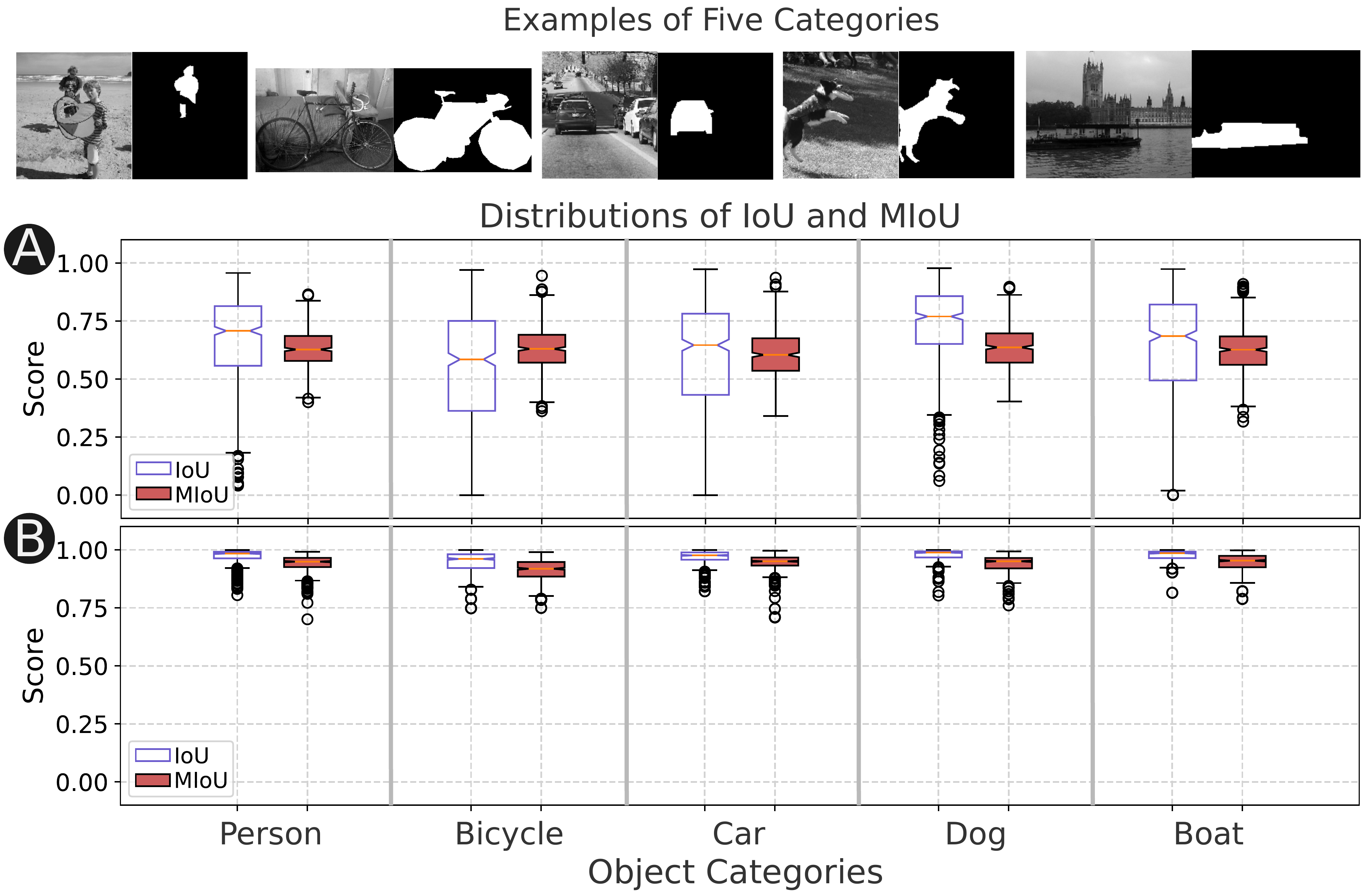}
        \caption{\footnotesize{Comparison of distributions of IoU and MIoU on 2500 masks obtained from five categories of COCO dataset.}}
        \label{fig:example_x}
    \end{figure}
    
    In this section, we present three experiments to verify the advantage of MIoU over IoU, and its reliability. The first experiment is conducted on a set of synthetic samples, and the two other ones utilize a larger set of everyday objects.

    \noindent\textbf{On Synthetic Regions.} In order to evaluate the sensitivity of MIoU, compared to other metrics, we need to control for the confounding factors, i.e., the random misalignments of the detected regions with respect to the ground-truth regions. Therefore, we generate sets of synthetic regions, as follows: each set corresponds to one ground-truth object, and contains several proposed regions and one region that perfectly aligns with the ground-truth region. For brevity, from several experiments that we ran, we present only one here. In this experiment, as illustrated in Fig. \ref{fig:example_f}, a table of 28 proposed regions are generated resulting from a systematic deviation from the ground-truth region, by means of linear scaling (mid-to-tail rows), translation (mid-to-head rows), and smoothing (left to right). The goal is to examine the sensitivity of IoU and MIoU to the jagged patterns, in the presence of different transformations. Comparing each of these regions with the ground-truth region, i.e., mask 3-1, the measured similarity is expected to decrease from left to right (in the table of samples), as the proposed regions lose their jagged structure. Following this expectation, a metric that is sensitive to the boundary structure of regions must show a periodically decreasing pattern, peaking at the beginning of each row. As the line plot shows, although all other metrics capture the between-row differences (periodically plateaued), only MIoU is sensitive enough to reflect the within-row differences, hence the periodically decreasing pattern.

    \noindent\textbf{On Real Regions.} To compare the distribution of MIoU with that of IoU, on real objects, we randomly sample a total of 2500 instances of COCO dataset \cite{lin2014microsoft}, equally collected from five categories (cars, bicycles, boats, dogs, and persons). We then apply some minimal manipulations to the ground-truth regions to generate two groups of proposed regions: The first group contains copies of the ground-truth regions which are randomly rotated ($\pm 10$ deg.) and/or translated ($\pm 10$ px). The second group is made of the smoothened duplicates of the ground-truth regions, using Gaussian smoothing followed by thresholding to obtain binary masks. In both experiments we set $\Delta = \{2^n, n < 10\}$. The box plots of the results are shown in Fig. \ref{fig:example_x}, where plot A corresponds to the first group, and plot B, to the second group. Overall, despite the fundamental differences between the two metrics' definitions, their reported quantities agree with each other. We observed this in several other experiments, which makes MIoU a reliable alternative. Moreover, in plot A, MIoU's distributions are centered at the median, across categories, with a significantly smaller variance than that of IoU. This is, we believe, the result of MIoU's intended sensitivity to the regions' boundary structure, rather than just the area of intersection. The pairs of distributions in plot B are also very similar while MIoU's median remains slightly below IoU's across all five categories. This consistent difference confirms the sensitivity of MIoU to the details that were smoothened out in this experiment.

%
%
\section{Conclusion}\label{sec:conclusion}
    For the general-purpose object-detection algorithms to be utilized on scientific computer vision tasks, fine segmentation of objects is needed. In this work, we highlighted the insensitivity of popular evaluation metrics to fine structure of the detected objects, and proposed a new metric to alleviate this issue. Our experiments showed that not only this is a reliable metric with a distribution consistent with IoU, but also exhibits sensitivity to the fine boundary structure of regions. We hope this study opens up the door for similar efforts to readjust our research horizon towards a smaller gap between the competitions' objectives and real-world challenges.

\section{Acknowledgments}
    This project has been supported in part by funding from CISE, MPS and GEO Directorates under NSF award \#1931555, and by funding from the LWS Program, under NASA award \#NNX15AF39G.




\bibliographystyle{IEEEbib}
\bibliography{strings,refs}

\begin{thebibliography}{10}

\bibitem{gavrila1999real}
Dariu~M Gavrila and Vasanth Philomin,
\newblock ``Real-time object detection for" smart" vehicles,''
\newblock in {\em Proceedings of the Seventh IEEE International Conference on
  Computer Vision}. IEEE, 1999, vol.~1, pp. 87--93.

\bibitem{karpathy2015deep}
Andrej Karpathy and Li~Fei-Fei,
\newblock ``Deep visual-semantic alignments for generating image
  descriptions,''
\newblock in {\em Proceedings of the IEEE conference on computer vision and
  pattern recognition}, 2015, pp. 3128--3137.

\bibitem{xu2015show}
Kelvin Xu, Jimmy Ba, Ryan Kiros, et~al.,
\newblock ``Show, attend and tell: Neural image caption generation with visual
  attention,''
\newblock in {\em International conference on machine learning}, 2015, pp.
  2048--2057.

\bibitem{bargoti2017deep}
Suchet Bargoti and James Underwood,
\newblock ``Deep fruit detection in orchards,''
\newblock in {\em 2017 IEEE International Conference on Robotics and Automation
  (ICRA)}. IEEE, 2017, pp. 3626--3633.

\bibitem{kupyn2018deblurgan}
Orest Kupyn, Volodymyr Budzan, Mykola Mykhailych, et~al.,
\newblock ``Deblurgan: Blind motion deblurring using conditional adversarial
  networks,''
\newblock in {\em Proceedings of the IEEE conference on computer vision and
  pattern recognition}, 2018, pp. 8183--8192.

\bibitem{ahmadzadeh2019toward}
Azim Ahmadzadeh, Sushant~S Mahajan, Dustin~J Kempton, et~al.,
\newblock ``Toward filament segmentation using deep neural networks,''
\newblock in {\em 2019 IEEE International Conference on Big Data (Big Data)}.
  IEEE, 2019, pp. 4932--4941.

\bibitem{lecun2019deep}
Y.~{LeCun},
\newblock ``1.1 deep learning hardware: Past, present, and future,''
\newblock in {\em 2019 IEEE International Solid- State Circuits Conference -
  (ISSCC)}, 2019, pp. 12--19.

\bibitem{krizhevsky2012imagenet}
Alex Krizhevsky, Ilya Sutskever, and Geoffrey~E Hinton,
\newblock ``Imagenet classification with deep convolutional neural networks,''
\newblock {\em Communications of the ACM}, vol. 60, no. 6, pp. 84--90, 2017.

\bibitem{ILSVRC15}
Olga Russakovsky, Jia Deng, Hao Su, et~al.,
\newblock ``{ImageNet Large Scale Visual Recognition Challenge},''
\newblock {\em International Journal of Computer Vision (IJCV)}, vol. 115, no.
  3, pp. 211--252, 2015.

\bibitem{alom2018history}
Md~Zahangir Alom, Tarek~M Taha, Christopher Yakopcic, et~al.,
\newblock ``The history began from alexnet: A comprehensive survey on deep
  learning approaches,''
\newblock {\em arXiv preprint arXiv:1803.01164}, 2018.

\bibitem{zou2019object}
Zhengxia Zou, Zhenwei Shi, Yuhong Guo, and Jieping Ye,
\newblock ``Object detection in 20 years: A survey,'' 2019.

\bibitem{zhao2019object}
Z.~{Zhao}, P.~{Zheng}, S.~{Xu}, and X.~{Wu},
\newblock ``Object detection with deep learning: A review,''
\newblock {\em IEEE Transactions on Neural Networks and Learning Systems}, vol.
  30, no. 11, pp. 3212--3232, 2019.

\bibitem{everingham2010pascal}
Mark Everingham, Luc Van~Gool, Christopher~KI Williams, et~al.,
\newblock ``The pascal visual object classes (voc) challenge,''
\newblock {\em International journal of computer vision}, vol. 88, no. 2, pp.
  303--338, 2010.

\bibitem{lin2014microsoft}
Tsung-Yi Lin, Michael Maire, Serge Belongie, et~al.,
\newblock ``Microsoft coco: Common objects in context,''
\newblock in {\em European conference on computer vision}. Springer, 2014, pp.
  740--755.

\bibitem{huang1995quantitative}
Qian Huang and Byron Dom,
\newblock ``Quantitative methods of evaluating image segmentation,''
\newblock in {\em Proceedings., International Conference on Image Processing}.
  IEEE, 1995, vol.~3, pp. 53--56.

\bibitem{estrada2009benchmarking}
Francisco~J Estrada and Allan~D Jepson,
\newblock ``Benchmarking image segmentation algorithms,''
\newblock {\em International Journal of Computer Vision}, vol. 85, no. 2, pp.
  167--181, 2009.

\bibitem{zou2004statistical}
Kelly~H Zou, Simon~K Warfield, Aditya Bharatha, et~al.,
\newblock ``Statistical validation of image segmentation quality based on a
  spatial overlap index1: scientific reports,''
\newblock {\em Academic radiology}, vol. 11, no. 2, pp. 178--189, 2004.

\bibitem{MartinFTM01}
D.~Martin, C.~Fowlkes, D.~Tal, and J.~Malik,
\newblock ``A database of human segmented natural images and its application to
  evaluating segmentation algorithms and measuring ecological statistics,''
\newblock in {\em Proc. 8th Int'l Conf. Computer Vision}, July 2001, vol.~2,
  pp. 416--423.

\bibitem{he2017mask}
Kaiming He, Georgia Gkioxari, Piotr Doll{\'a}r, and Ross Girshick,
\newblock ``Mask r-cnn,''
\newblock in {\em Proceedings of the IEEE international conference on computer
  vision}, 2017, pp. 2961--2969.

\bibitem{keogh2006lb_keogh}
Eamonn Keogh, Li~Wei, Xiaopeng Xi, et~al.,
\newblock ``Lb\_keogh supports exact indexing of shapes under rotation
  invariance with arbitrary representations and distance measures,''
\newblock in {\em Proceedings of the 32nd international conference on Very
  large data bases}. VLDB Endowment, 2006, pp. 882--893.

\bibitem{denker1999synoptic}
C~Denker, A~Johannesson, W~Marquette, et~al.,
\newblock ``Synoptic h$\alpha$ full-disk observations of the sun from big bear
  solar observatory--i. instrumentation, image processing, data products, and
  first results,''
\newblock {\em Solar Physics}, vol. 184, no. 1, pp. 87--102, 1999.

\bibitem{kumar2012leafsnap}
Neeraj Kumar, Peter~N Belhumeur, Arijit Biswas, et~al.,
\newblock ``Leafsnap: A computer vision system for automatic plant species
  identification,''
\newblock in {\em European conference on computer vision}. Springer, 2012, pp.
  502--516.

\bibitem{theiler1990estimating}
James {Theiler},
\newblock ``Estimating fractal dimension,''
\newblock {\em Journal of The Optical Society of America A-optics Image Science
  and Vision}, vol. 7, no. 6, pp. 1055--1073, 1990.

\bibitem{mandelbrot1982fractal}
Benoit~B Mandelbrot,
\newblock ``The fractal geometry of nature. 1982,''
\newblock {\em San Francisco, CA}, 1982.

\bibitem{barnsley1988science}
Michael~F Barnsley, Robert~L Devaney, Benoit~B Mandelbrot, et~al.,
\newblock {\em The science of fractal images},
\newblock Springer, 1988.

\end{thebibliography}
\balance

\end{document}